\documentclass[conference]{IEEEtran}
\usepackage{cite}
\usepackage{amsmath,amssymb,amsfonts}
\usepackage{algorithmic}
\usepackage{graphicx}
\usepackage{textcomp}
\usepackage{xcolor}
\usepackage{booktabs}
\usepackage{multirow}
\usepackage{array}
\usepackage{float}  

\def\BibTeX{{\rm B\kern-.05em{\sc i\kern-.025em b}\kern-.08em
    T\kern-.1667em\lower.7ex\hbox{E}\kern-.125emX}}
    
\begin{document}

\title{Prediction of Highway Traffic Flow Based on Artificial Intelligence Algorithms Using California Traffic Data}

\author{
\IEEEauthorblockN{Junseong Lee\IEEEauthorrefmark{1}, Jaegwan Cho\IEEEauthorrefmark{1}, Yoonju Cho\IEEEauthorrefmark{1}, Seoyoon Choi\IEEEauthorrefmark{2}, Yejin Shin\IEEEauthorrefmark{2}}
\IEEEauthorblockA{\IEEEauthorrefmark{1}Yonsei University, \IEEEauthorrefmark{2}Sookmyung Women's University\\
brulee@yonsei.ac.kr, jjgs1235@yonsei.ac.kr, jessicaiq10@yonsei.ac.kr, sseo9872@gmail.com, yejins0605@gmail.com}
}

\maketitle

\begin{abstract}
The study "Prediction of Highway Traffic Flow Based on Artificial Intelligence Algorithms Using California Traffic Data" presents a machine learning-based traffic flow prediction model to address global traffic congestion issues. The research utilized 30-second interval traffic data from California Highway 78 over a five-month period from July to November 2022, analyzing a 7.24km westbound section connecting "Melrose Dr" and "El-Camino Real" in the San Diego area. The study employed Multiple Linear Regression (MLR) and Random Forest (RF) algorithms, analyzing data collection intervals ranging from 30 seconds to 15 minutes. Using R², MAE, and RMSE as performance metrics, the analysis revealed that both MLR and RF models performed optimally with 10-minute data collection intervals. These findings are expected to contribute to future traffic congestion solutions and efficient traffic management.
\end{abstract}

\section{Introduction}

Currently, traffic congestion is one of the most pressing issues faced globally. According to the California Air Resources Board, a decrease of 10 miles per hour in speed due to congestion results in an increase of approximately 100 grams of CO$_2$ emissions per mile \cite{carb2017}. The 2023 Global Traffic Scorecard report by the traffic analytics company INRIX states that the average American driver loses 42 hours annually—equivalent to an entire workweek—due to traffic congestion, resulting in a time loss valued at \cite{cnbc2024}.

Kang et al. utilized LSTM to predict traffic volume at a target detector location, experimenting with varying the number of upstream and downstream detectors used as inputs \cite{kang2017}. Nevertheless, their experiment was conducted using 17,557 sample points, focusing on a broad area rather than on downstream traffic volume prediction for a single road segment. M. S. A. Siddiquee et al. predicted daily traffic volume on Bangladeshi highways using hourly traffic data. By employing an artificial neural network (ANN), they trained on intermittent data to estimate missing values and predict not only daily but also hourly traffic volumes \cite{siddiquee2017}.

The structure of this paper is as follows. Section II describes the original dataset used in this study and the preprocessing steps taken. It also presents the final dataset generated through preprocessing. Section III outlines the AI algorithms and performance metrics employed in the study. Section IV compares and analyzes the performance results of the AI-based models considered. Finally, Section V discusses the conclusions drawn from this study and suggests directions for future research.

\section{Dataset Configuration and Data Preprocessing}

\subsection{Dataset Description}

To analyze traffic flow in the San Diego area, data from California State Route 78 was utilized. The study section covers approximately 7.24 km of westbound roadway, connecting "Melrose Dr" and "El-Camino Real." The data was collected over a five-month period during the second half of 2022, specifically from July to November, with measurements recorded every 30 seconds around the clock. We received the dataset from Caltrans: California Department of Transportation for research purposes only.

\begin{figure}[!t]
\centering
\includegraphics[width=\columnwidth]{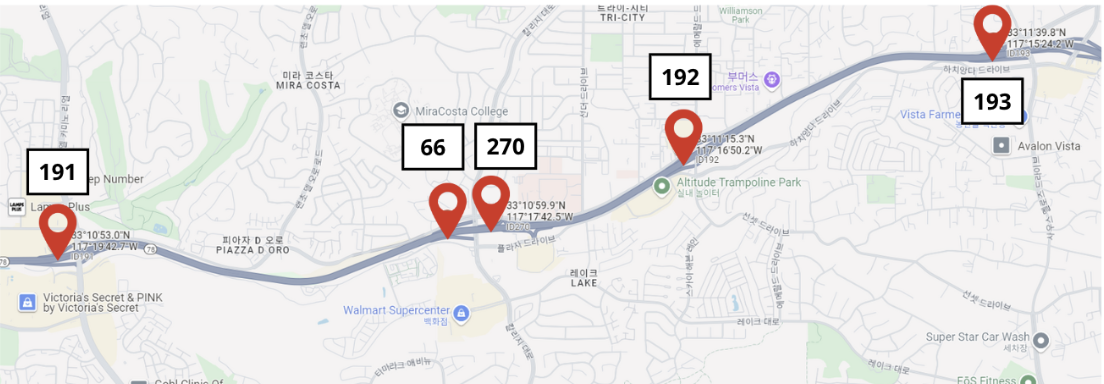}
\caption{State Route 78 in San Diego County (Traffic Flow: Westbound).}
\label{fig:route78}
\end{figure}

Figure \ref{fig:route78} shows the locations of loop detectors installed to measure traffic volume, which are distributed across both the mainline and the on/off ramps. Using this equipment, the California Department of Transportation records the number of passing vehicles and roadway occupancy every 30 seconds. Operating at 30Hz, this system can collect up to 900 data points within each 30-second interval.

Table \ref{tab:variables} describes the attributes that make up the traffic dataset for State Route 78 in San Diego County. The dataset includes the following elements: measurement date (Date), end time of each 30-second measurement interval (Time), detector identification number (ID), name of the highway entry/exit location (Name), number of vehicles passing through each mainline lane during each 30-second interval (LaneX\_Vol), occupancy for those vehicles (LaneX\_Occ). The original Caltrans dataset used in this study contains measurements at 30-second intervals, totaling 426,240 data samples.

\begin{table}[!t]
\centering
\caption{Variable Names and Characteristics of the Original Caltrans Dataset}
\label{tab:variables}
\begin{tabular}{|l|l|l|}
\hline
\textbf{Variables} & \textbf{Type} & \textbf{Unit} \\
\hline
Date & Year-Month-Day & 2022.07.01-2022.11.30 \\
\hline
Time & Hour-Minute-Second & 00:00:00-23:59:30 \\
\hline
ID & Integer & 66,191,192,193,270 \\
\hline
LaneX\_Vol & Integer & 0,1,... \\
\hline
LaneX\_Occ & Integer & 0,1,... \\
\hline
OffX\_cnt & Integer & 0,1,... \\
\hline
PsgX\_cnt & Integer & 0,1,... \\
\hline
\end{tabular}
\end{table}

Figure \ref{fig:route78} illustrates the section of State Route 78 in San Diego County, California, considered in this study. A total of five loop detectors are installed along the mainline of the highway. The corresponding entry and exit road names for detectors ID 191, ID 66, ID 270, ID 192, and ID 193 are El-Camino Real, College Blvd., Emerald Dr., and again El-Camino Real, respectively.

Among these, detector ID 193 is located at the most upstream point, while ID 191 is positioned at the most downstream point. The objective of this study is to predict the traffic volume at ID 191 by utilizing the volume and occupancy data of ID 191 itself with various historical lengths.

Among the total 426,240 data samples, 1.14\% contained missing values, which were supplemented using linear interpolation.

\subsection{Data Preprocessing Procedure}

The data preprocessing in this study was conducted in three main steps: data integration, time unit adjustment, and target data selection for analysis.

First, during the raw data integration step, data scattered across individual detectors were reorganized. Specifically, the traffic volume and occupancy data for each lane (Lane 1, 2, 3) were aggregated into single values. The data was then categorized by detector ID and transformed into a pivot table format based on date and time.

Second, to enable analysis at various time resolutions, the original 30-second (0.5-minute) interval data was restructured into 1-minute, 2-minute, 5-minute, 10-minute, and 15-minute intervals. The reconstruction followed this rule: when creating higher time-unit datasets, all traffic volume and occupancy values within the target time range were summed, while the month and time information were taken from the first data entry of each interval. For example, 1-minute interval data was generated by combining two consecutive 30-second records, and 2-minute data by combining four consecutive 30-second records.

Lastly, to improve analytical accuracy, a weekday-based data selection process was performed. Based on the date information, a weekday index (1–7, Sunday–Saturday) was assigned to each data sample, after which the data was divided into weekdays (2–6) and weekends (1, 7). To ensure consistency in traffic patterns, weekend data was excluded from the analysis.

Through this preprocessing, a structured weekday traffic dataset was established for each time unit, serving as the foundational resource for subsequent analysis.

\begin{figure}[!t]
\centering
\includegraphics[width=\columnwidth]{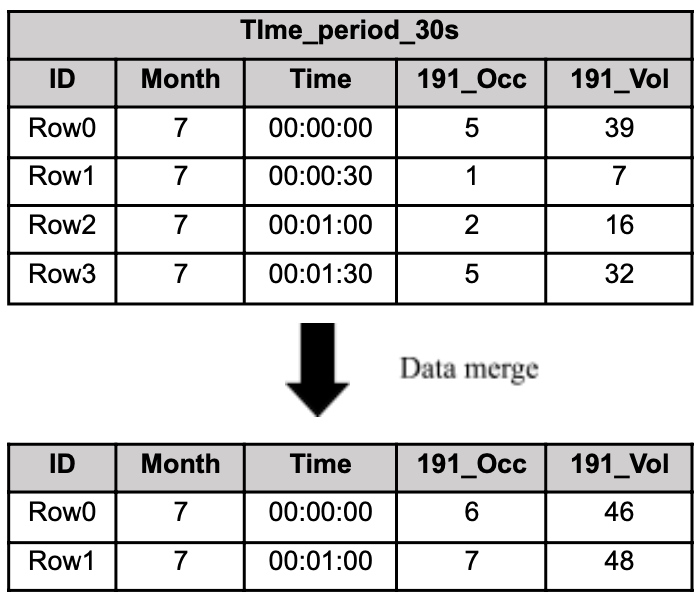}
\caption{Time Unit Adjustment Process.}
\label{fig:time_adjustment}
\end{figure}

Figure \ref{fig:time_adjustment} presents a table illustrating an example of the second step in the three-stage data preprocessing procedure: time unit adjustment. Through these three steps, the final dataset, referred to as the CFD, was constructed. The results are summarized below.

\begin{table}[H]
\centering
\caption{Final Constructed Feature Dataset (CFD\_T)}
\label{tab:final_dataset}
\begin{tabular}{|l|l|l|l|}
\hline
\textbf{Variables} & \textbf{Type} & \textbf{Values} & \textbf{Unit} \\
\hline
\multicolumn{4}{|c|}{\textbf{Independent Variable}} \\
\hline
Month & Integer & \{7, ..., 11\} & - \\
\hline
Time & Double & \{-1.0, ..., 1.0\} & - \\
\hline
191\_Occ\_T & Double & \{0.0, ...\} & - \\
\hline
\multicolumn{4}{|c|}{\textbf{Dependent Variable}} \\
\hline
191\_Vol\_T & Double & \{0.0, ...\} & - \\
\hline
\end{tabular}
\end{table}

Here, T = \{0.5 min, 1 min, 2 min, 5 min, 10 min, 15 min\}.

\section{Machine Learning Algorithms and Performance Evaluation Metrics}

In this study, Multiple Linear Regression (MLR) and Random Forest (RF) algorithms were employed to predict traffic volume. The performance of the traffic volume prediction models was compared and analyzed using evaluation metrics.

\subsection{Multiple Linear Regression (MLR)}

Multiple Linear Regression is a statistical technique that uses multiple input variables (independent variables) to predict a single output variable (dependent variable). Each independent variable is assigned a weight (regression coefficient), and the final predicted value is calculated by summing these weighted inputs. The model takes the form \( Y = \beta_0 + \beta_1 X_1 + \beta_2 X_2 + \dots + \beta X \), with the objective of finding the set of \(\beta\) values that best explains the observed data.

\subsection{Random Forest (RF)}

Random Forest is an ensemble machine learning algorithm that combines multiple decision trees to perform accurate predictions for classification tasks. It offers stable prediction performance across various datasets and is particularly effective with complex datasets that exhibit nonlinear relationships. Although RF-based models can suffer from overfitting, this study mitigated that risk by employing early stopping techniques to determine optimal hyperparameters. Specifically, tree depth and the minimum number of data samples required at leaf nodes were considered as key hyperparameters.

\subsection{Performance Metrics}

\textbf{R² (R-squared):} This metric indicates how well the regression model explains the variance of the data. It represents the proportion of total variance explained by the model. A value closer to 1 suggests that the model captures the data patterns effectively, though the possibility of overfitting should also be considered.

\textbf{Mean Absolute Error (MAE):} This is the average of the absolute differences between the predicted and actual values. It is intuitive and less sensitive to outliers compared to other metrics. MAE shows how much the predictions deviate from the actual values on average, expressed in the original unit.

\textbf{Root Mean Squared Error (RMSE):} This is the square root of the average of the squared differences between predicted and actual values. It places more weight on larger errors, making it more sensitive to significant prediction errors. Like MAE, RMSE is interpretable in the original unit.

\section{Traffic Volume Prediction Model Analysis}

\subsection{Model Training and Hyperparameter Settings}

This study used a total of two machine learning algorithms. The training and test datasets were split at an 80\% to 20\% ratio using linear sampling. For Random Forest, two hyperparameters were configured: minimum child node and maximum tree depth. The optimization range for these parameters was set from 3 to 25 and from 2 to 20, respectively. The optimal hyperparameters were selected based on the results from a total of 500 decision trees.

\subsection{Traffic Volume Prediction Model Results}

\begin{figure}[H]
\centering
\includegraphics[width=\columnwidth]{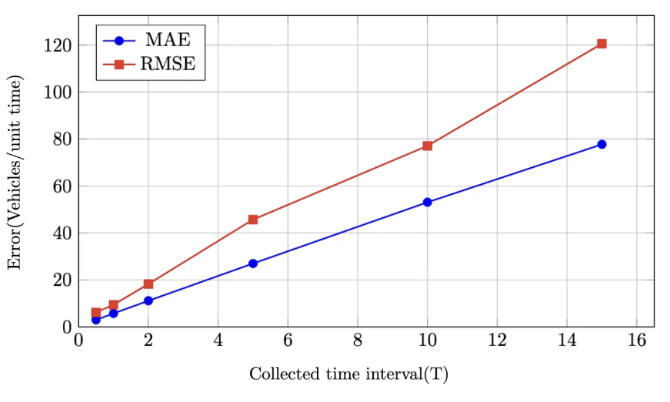}
\caption{Comparison of MLR Model Performance (MAE, RMSE) for the CFD\_T Dataset.}
\label{fig:mlr_mae_rmse}
\end{figure}

Figure \ref{fig:mlr_mae_rmse} shows the performance of the MLR model trained on the CFD\_T dataset. The study examined how the model performed with varying data collection intervals ranging from 30 seconds to 15 minutes, as described in Section 2.2.

As the data collection interval increased, both MAE and RMSE showed a steady upward trend. Additionally, R² increased up to the 10-minute interval but began to decline when the interval reached 15 minutes.

\begin{figure}[H]
\centering
\includegraphics[width=\columnwidth]{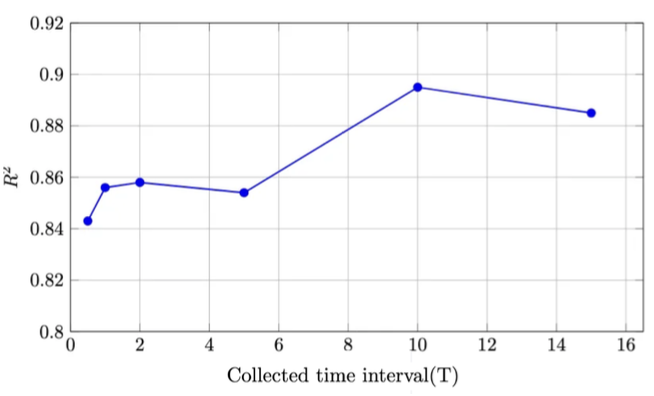}
\caption{Comparison of MLR Model Performance (R²) for the CFD\_T Dataset.}
\label{fig:mlr_r2}
\end{figure}

However, if the collected time intervals are different, direct comparison with the original Caltrans data collected at 30-second intervals is not possible. Therefore, we compared performance using Scaled MAE/RMSE. Here, Scaled error is defined as:

\begin{equation}
Error_{scaled} = \frac{Error_{origin} \times 0.5min}{T}
\end{equation}

where $T = \{0.5, 1, 2, 5, 10, 15\}$ min.

\begin{figure}[H]
\centering
\includegraphics[width=\columnwidth]{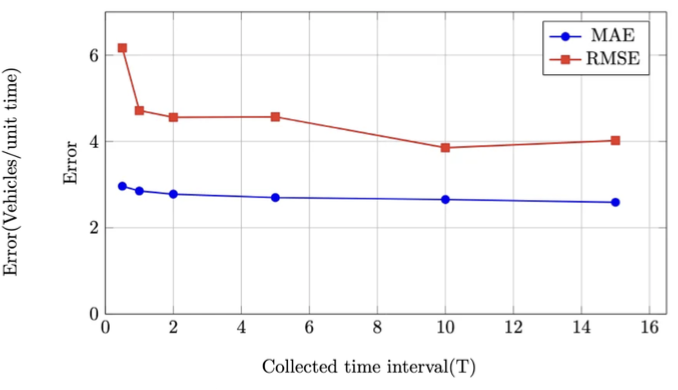}
\caption{Comparison of MLR Model Scaled Performance (MAE, RMSE) for the CFD\_T Dataset.}
\label{fig:mlr_scaled}
\end{figure}

Finally, from Figure \ref{fig:mlr_scaled}, it was observed that the Scaled MAE decreased up to a collection interval of 10 minutes and then showed an increasing trend after the 15-minute collection interval. Additionally, the performance when applying the same CFD\_T dataset to the Random Forest algorithm was as follows.

\begin{figure}[!h]
\centering
\includegraphics[width=\columnwidth]{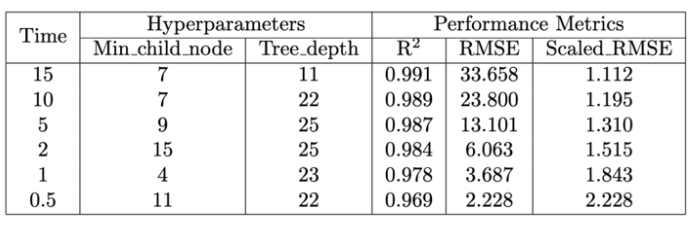}
\caption{Comparison of RF Model Performance (MAE, RMSE) for the CFD\_T Dataset.}
\label{fig:rf_mae_rmse}
\end{figure}

For the RF model, as you can see below, 
we also confirmed the trend and observed that performance improved up to a 15-minute data collection interval.

\begin{figure}[H]
\centering
\includegraphics[width=\columnwidth]{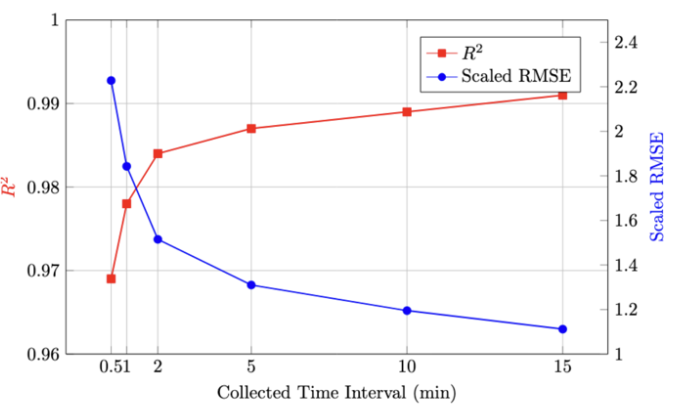}
\caption{Comparison of RF Model Performance (R²) for the CFD\_T Dataset.}
\label{fig:rf_r2}
\end{figure}

\section{Conclusion}

In this study, we proposed machine learning-based traffic prediction models utilizing data from various time intervals to improve traffic prediction accuracy. The machine learning algorithms considered were Multiple Linear Regression (MLR) and Random Forest (RF). R², MAE, and RMSE were used as performance metrics to evaluate the traffic prediction models. Additionally, scaled errors were employed to evaluate RMSE values from different time interval datasets on a consistent basis.

MLR analysis showed performance improvements up to 10-minute collection intervals, with a noticeable degradation in performance at 15-minute intervals. Based on this, we determined that 10 minutes is the optimal data collection interval for MLR. Furthermore, in the RF experiments, performance continued to improve up to 15-minute collection intervals.

For future research, our goal is to meticulously analyze highway traffic volumes using not only the datasets constructed in this study but also a greater number of detector IDs. We also aim to analyze the optimal time interval for the RF model by exploring even longer collection time intervals.

\section*{Acknowledgment}

This research was supported by the MSIT(Ministry of Science and ICT), Korea, under the National Program for Excellence in SW(2023-0-00054) supervised by the IITP(Institute of Information \& communications Technology Planning \& Evaluation) in 2025. Also, This research was supported by the MSIT (Ministry of Science, ICT), Korea, under the National Program for Excellence in SW), supervised by the IITP(Institute of Information \& communications Technology Planning \& Evaluation) in 2025(2022-0-01087) \& (2024-0-00062).

\end{document}